\pdfoutput=1

\documentclass[final,3p,times]{elsarticle}

\usepackage{amssymb}
\usepackage{amsthm}
\usepackage{amsmath}
\usepackage{lipsum}
\usepackage{multirow}
\usepackage{color}
\usepackage{graphics}
\usepackage{ifpdf}

\usepackage{tablefootnote}

\journal{Neurocomputing}
\newcommand{\KHR}{\textcolor{black}}
\begin{document}

\begin{frontmatter}



\title{Micro-expression spotting: A new benchmark}


\author[oulu]{Thuong-Khanh~Tran}
\author[oulu]{Quang-Nhat~Vo}        
\author[xian]{Xiaopeng~Hong}      
\author[oulu]{Xiaobai~Li}      
\author[oulu]{Guoying~Zhao}      

\address[oulu]{Center for Machine Vision and Signal Analysis, University of Oulu, Finland}
\address[xian]{School of Electronic and Information Engineering, Xi'an Jiaotong University, China}

\begin{abstract}
Micro-expressions (MEs) are brief and involuntary facial expressions that occur when people are trying to hide their true feelings or conceal their emotions. Based on psychology research, MEs play an important role in understanding genuine emotions, which leads to many potential applications. Therefore, ME analysis has become an attractive topic for various research areas, such as psychology, law enforcement, and psychotherapy.  In the computer vision field, the study of MEs can be divided into two main tasks, spotting and recognition, which are used to identify positions of MEs in videos and determine the emotion category of the detected MEs, respectively. 
Recently, although much research has been done, no fully automatic system for analyzing MEs has yet been constructed on a practical level for two main reasons: most of the research on MEs only focuses on the recognition part, while abandoning the spotting task; current public datasets for ME spotting are not challenging enough to support developing a robust spotting algorithm.
The contributions of this paper are threefold: (1) we introduce an extension of the SMIC-E database, namely the SMIC-E-Long database, which is a new challenging benchmark for ME spotting;
(2) we suggest a new evaluation protocol that standardizes the comparison of various ME spotting techniques;
(3) extensive experiments with handcrafted and deep learning-based approaches on the SMIC-E-Long database are performed for baseline evaluation.
\end{abstract}




\end{frontmatter}

\section{Introduction}
\label{introduction}
Affective computing is a research field that processes, recognizes, interprets, and simulates human emotions, which play an important role in human-machine interaction analysis. Affective computing can be related to voices, facial expressions, gestures, and bio-signals \cite{zeng2008survey}. Among them, facial expressions (FEs) are certainly one of the most important channels used by people to convey internal emotions. There has been much research on the topic of FE recognition. Several state-of-the-art FE recognition methods reported an accuracy rate of more than 90\% [1]. Aside from ordinary FEs, under certain cases emotions can manifest themselves in a special form called  ``micro-expressions'' (MEs). \cite{li2017towards,oh2018survey,wu2011machine}.

MEs are brief and involuntary facial expressions that occur when people are trying to hide their true feelings or conceal their emotions \cite{li2015reading}. Research from Ekman \cite{ekman2009lie} shows that MEs play an important role in psychology by helping us understand hidden emotions. Spontaneous micro-expressions may occur swiftly and involuntarily, and they are difficult to actively and willingly control. This characteristic of MEs makes several applications possible. For example, when evaluating the performance of lectures based on a student’s emotions, a fleeting ME can reveal the
normally hidden emotions of students. In addition to its potential applications in education, ME analysis has promise for other fields as well, such as medicine, business, and national security \cite{li2017towards}. In business, a salesperson can use MEs to determine a customer’s true response when introducing new products. Border control agents can detect abnormal behaviors when asking people questions. For example, the U.S. Transportation Security Administration has developed the SPOT program, in which airport staff are trained to observe passengers displaying suspicious behaviors by analyzing MEs and conversational signs \cite{li2015reading}. In medical fields, especially in psychotherapy, doctors can use MEs as a clue for understanding a patient’s genuine feelings \cite{li2017towards}.
Therefore, ME analysis plays an essential role in analyzing people’s hidden feelings in various contexts.

Unlike regular facial expressions, which we can recognize effortlessly, reading MEs is very difficult for humans because they often occur too briefly and quickly for human eyes to spot and recognize. ME recognition by humans is still deficient, even when done by a well-trained specialist. In contrast, many studies in the computer vision field have reported the impressive performance of Facial Expression Recognition and Facial Analysis tasks \cite{li2018deep}. Consequently, it is important to assess how best to utilize computer science, especially computer vision, for ME analysis.
 
Research on automatic ME analysis in computer vision has addressed to two main problems: recognition and spotting. The former has to do with determining the emotional state of MEs, while the latter is about locating the temporal positions of MEs in video sequences. Currently, most ME research focuses on recognition tasks via many novel techniques   \cite{oh2018survey,van2019capsulenet}, while ME spotting still has yielded only limited results. Detecting ME is probably even more challenging than ME recognition. One study \cite{li2017towards} has indirectly referred to this issue in human test experiments, where automatic ME recognition methods outperformed humans, while the accuracy rate has proved much lower when the ME spotting task is added to the ME analysis system. For real-world applications, ME positions must be determined first before any further emotion recognition or interpretative operations can be performed. Therefore, ME spotting is an important task in developing a fully automatic ME analysis system.
 
Although increasing attention has been paid to the spotting task, several issues remain unresolved. First, the database for ME spotting is quite limited. According to one survey  \cite{oh2018survey}, there are only four spontaneous databases suitable to evaluate spotting methods. Additionally, the existing ME spotting studies have been evaluated based on a limited number of databases. The length of videos in several datasets are often just 5–10 seconds and do not consist of challenging cases: eye blinking, head movements, and regular facial expressions, which are easily confused with MEs. If we consider building a real environment system for ME analysis, we need to evaluate the ME spotting methods on longer videos with more complex facial behaviors. Thus, creating more challenging databases for ME spotting is an important task. Second, it is difficult to adequately compare existing techniques because they were often evaluated with
different evaluation protocols. Moreover, existing evaluation protocols only focus on the correct location of spotted ME samples without considering the accuracy of detected ME intervals. Several recognition techniques require the correct location of onset/apex/offset frames for recognition  \cite{van2019capsulenet, oh2018survey, li2018can,xia2019spatiotemporal}. Recently, most ME recognition studies have been developed and tested on manually segmented ME samples with a correct border, so the accuracy would largely be reduced if the inputted ME samples contain incorrect frames. For example, Li et al. \cite{li2017towards} found that the performance of ME recognition is greater than $65\%$ when evaluated using manually processed ME samples. However, when the authors evaluated the entire ME analysis system, which consisted of both spotting and recognition, the recognition performance proved less than  $55\%$. 
Thus, if the returned ME samples have many neutral frames or incorrect locations in the apex frames, it can diminish the performance of the recognition task. Consequently, it is meaningful to standardize the performance evaluation of ME spotting when conducting experiments in the same evaluation setting and using the same metric.

To overcome the above-mentioned issues, our study makes three important contributions:
 \begin{itemize}
 \item We introduce a new challenging database for ME spotting, which is the new benchmark for the spotting task.  
 \item We suggest a new set of protocols that standardize the evaluation of ME spotting methods. Our protocols aim to make a fair comparison among spotting techniques by considering both the correct ME locations and ME intervals, which are two important factors for the subsequent ME recognition step. 
 \item With the new ME database and proposed evaluation protocols, we evaluate several recent ME spotting approaches, including traditional approaches and deep learning approaches, and provide baseline results for reference in future ME spotting studies.
 \end{itemize}
 
The rest of this paper is organized as follows. The second section reviews works on ME spotting and relevant databases. The third section introduces a new ME spotting database. Next, we describe the selected spotting methods, which are used for providing the baseline results. Then, the proposed protocols are introduced in section \ref{evaluation}. Finally, the last section reports experimental results and presents our conclusions. 

\footnotetext{All the baseline source codes and the dataset will be released upon the publication of this work.}

\section{Related Work}
MEs include subtle facial emotions and low-intensity movements. Thus, they are difficult for ordinary people to spot. The spotting of MEs usually requires well-trained experts. 
To develop an automatic ME analysis system, several public ME databases and ME spotting studies have been proposed in the literature.

\subsection{Micro-expression database} 
\label{related_db}

\begin{figure*}[ht]
    \centering
  \includegraphics[width=15cm,height=8cm]{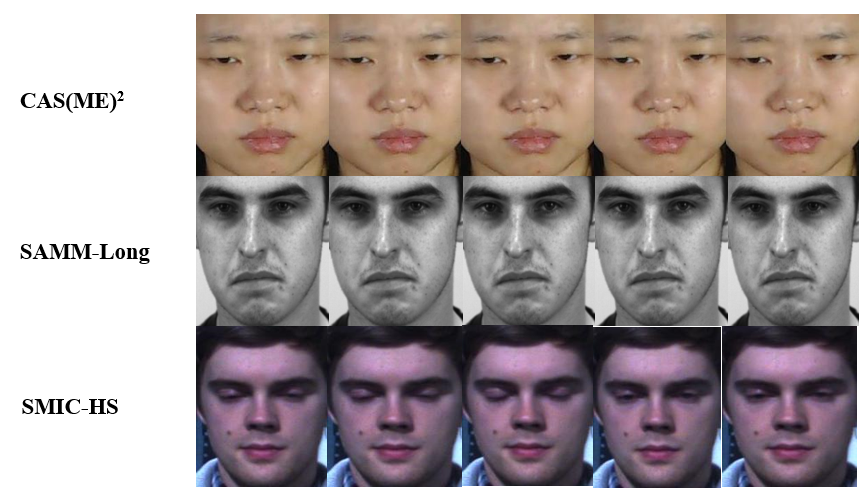}
  \caption{Examples of three ME databases for spotting. (Top) One example of  CAS(ME)$^{2}$ displays  ``anger'' from subject $15$. (Center) An example from the SAMM dataset shows the ME sample of ``sadness'' from subject $6$. (Bottom) ``Positive'' ME sample from subject $3$ of the SMIC-E dataset. }
  \label{fig:database}
\end{figure*}

Though ME analysis research in the computer vision field has become an attractive topic in the last few years, the number of public  ME databases is still limited. A survey done by Oh et al. \cite{oh2018survey} identified only 11 databases for ME research, with most of them only dealing with ME recognition. 
Additionally, several databases for ME spotting contain only the acted MEs, which are different from spontaneous ones; thus, the models built based on acted MEs would not work well in real-world applications. In the case of MEs, spontaneous emotions are more genuine than acted emotions. Therefore, we only summarize the spontaneous databases suitable for ME spotting.  
Figure \ref{fig:database} provides three examples of ME samples from the spontaneous databases. 
\KHR{Table \ref{tab:table_compare}, in turn, summarizes the parameters of existing ME spotting datasets.}

SMIC is the first database recorded by three different types of camera: high speed (SMIC-HS), normal visual (SMIC-VIS), and near-infrared (SMIC-NIS). In 2013, Li et al. \cite{smicdata} also extended the SMIC by adding non-micro frames before and after the labeled micro frames to create an extended version of SMIC, namely SMIC-HS-E, SMIC-VIS-E, and SMIC-NIR-E. This dataset contains 166 ME samples from 16 subjects. 

\KHR{
In 2013 and 2014, CASME and CASME-II were published for future ME analysis. In these two datasets, video samples have neutral frames after and before the ME frames and the annotation of onset/apex/offset can be utilized to evaluate the spotting performance \cite{li2016spontaneous}. However, CASME and CASME-II have a drawback: the average duration of video samples is only two or three seconds. Therefore, they are not suitable for ME spotting.
}

\KHR{Recently, a new dataset, namely CAS(ME)$^{2}$, was introduced for both expression recognition and spotting tasks \cite{casme2}. In this database, facial expressions consist of both macro-expressions and micro-expressions, with it being challenging for ME spotting techniques to discriminate between normal facial expressions and ME samples.
The CAS(ME)$^{2}$ database is divided into two parts: part A contains both spontaneous macro-expressions and MEs in long videos, and part B includes cropped expression samples with frames ranging from onset to offset. 
With respect to ME spotting, only part A is suitable for ME spotting. CAS(ME)$^{2}$ has 97 long videos from 22 subjects}.

\KHR{In the SAMM database \cite{davison2016samm}, the author published the newest version of SAMM-long, which is suitable for ME spotting. This dataset also contains both macro-expressions and micro-expressions. SAMM-long contains 147 videos from 32 subjects.
}

The recent increasing need for data acquired from unconstrained ``in-the-wild'' situations has prompted further efforts to provide more naturalistic, high-stake scenarios. The MEVIEW dataset \cite{husak2017spotting} was constructed by collecting poker game videos downloaded from YouTube with a close-up of the player's face. In total, MEVIEW consists of 41 videos of 16 subjects.

\KHR{Overall, most of the databases have been extended from existing ones by adding nature frames to after and before ME samples to create long videos. This approach is reasonable because recording new videos takes much effort and requires experts to label the ground truth. However, the effort to compile ME spotting databases still suffers from the fact that only two suitable ME spotting datasets exist, meaning that a real-life ME spotting system cannot yet be developed. Hence, a new ME spotting database with long videos and challenging cases of facial behaviors is needed.}

\subsection{Micro-expression spotting} 
\label{related_method}
When a potential application of ME analysis is implemented in real life, it needs to detect the temporal locations of ME events before any recognition step can be applied. Therefore, ME spotting is an indispensable module for a fully automated ME analysis system. Several studies have addressed this topic. In the scope of this paper, we will explore the current trends in ME spotting research, as discussed in the literature.

At first, almost all methods try to detect ME in videos by computing the feature difference between frames. For example, Moilanen et al. \cite{moilanen2014spotting} spotted MEs using the Chi-Square distance of the Local Binary Pattern (LBP) in fixed-length scanning windows. This method provides the baseline results in the first whole ME analysis system, which combines spotting and recognition  \cite{li2015reading}. Patel et al. \cite{patel2015spatiotemporal} proposed calculating optical flow vectors for small, local spatial regions, then using heuristic algorithms to remove non-MEs. Wang et al. \cite{wang2016main} suggested a method called Main Directional Maximal Differences, which utilizes the magnitude of maximal difference in the main direction of optical flow. Recently, Riesz transform combining with facial maps has been employed to spot MEs automatically \cite{duque2018micro}. 
Kai et al. \cite{beh2019micro} proposed spotting the spontaneous MEs in long videos by detecting the changes in the ratio of the Euclidean distances between facial landmarks in three facial regions. We categorize the above-mentioned techniques as the \textbf{unsupervised learning} approach. 

Nonetheless, tiny motions on the face, such as eye blinks and head movements, are usually very difficult to determine from ME samples using the thresholding techniques. Hence, some later works proposed utilizing machine learning techniques as a robust tool to distinguish between ME samples and normal facial behaviors. We categorize these methods as the \textbf{supervised learning} approach.
One research study \cite{xia2016spontaneous} has introduced the first attempt at utilizing machine learning in ME spotting. The authors employed Adaboost to estimate the probability of consecutive frames belonging to an ME. Then, they used random walk functions to refine and integrate the output from Adaboost to return the final result. In a more recent effort at providing a benchmark to standardize the evaluation measures in ME spotting studies, Tran et al. \cite{tran2017sliding} proposed using a multi-scale, sliding-window-based method to detect MEs. This method tackles ME spotting as a binary classification problem based on a window sliding across positions and scales of a video sequence. Although other studies \cite{xia2016spontaneous,tran2017sliding} take advantage of machine learning, the performances are still not good enough because traditional learning methods are still not robust enough to handle the subtle movements of MEs. 

Recently, deep learning has become a new trend in many computer vision fields that overcomes the limitations of traditional approaches. Several studies have applied deep learning to the ME spotting problem. Zhang et al. \cite{zhang2018smeconvnet} firstly proposed using a Convolutional Neural Network (CNN) to detect the apex frame in two main steps: (1) building CNN architectures to classify neutral frames and apex frames; (2) introducing a feature engineering method to merge nearby detected samples. However, since this method only trains the single-image classifier, it might raise false alarms in videos containing macro facial expressions. Tran et al. \cite{tran2019} proposed another deep learning-based technique by combining spatial-temporal features with deep sequence model to compute the apex score in longer videos. Their framework consists of two main steps: 1) based on each position of a video, they extracted a spatial-temporal feature that can identify MEs from amongst extrinsic movements; 2) they utilized an LSTM network for both local and global correlations of the extracted features to predict the ME apex frame score. However, this approach is only evaluated on short clips of ME datasets.

Although different techniques have been introduced for ME spotting and some progress has been made, the ability to apply them to a real ME analysis system has yet to be realized. One of the reasons for this limitation is that the ME databases currently used to develop ME spotting methods are still not challenging enough compared to real-life situations. For real-life applications, we need to evaluate ME spotters on longer and more challenging videos.

\subsection{Evaluation protocols} \label{related_protocol}
Together with the significantly increasing number of recent ME spotting studies, various evaluation metrics for ME spotting methods have likewise been proposed. A survey by Oh \cite{oh2018survey} claims that the evaluation of ME spotting can be divided into two main approaches: ME sample spotting-based evaluations and apex-frame spotting-based evaluations. 

Li et at \cite{li2015reading} propose considering the spotted apex frame inside the interval  [ onset $- \frac{N-1}{4},$ offset $+ \frac{N-1}{4}$] as the true positive, where $N$ is the length of detected window. They then plotted the Receiver Operating Characteristic (ROC) curve to evaluate their methods. 
Duque et al. \cite{duque2018micro} then built on the ROC by adding the Area Under Curve (AUC) as the standard evaluation metric.

In a follow-up study \cite{xia2016spontaneous}, the authors utilized the ROC curve to test their method, with the difference being that the true positive samples are detected by the Intersection over Union (IoU) between the detected sample and ground truth. This method is formulated as follows:

\begin{equation}
    \label{overlapposneg}
    \begin{split}
    IoU = \frac{X_{G} \cap  X_{W}}{X_{G} \cup  X_{W}}
    \\
    true\:positive: IoU \geq \varepsilon
    \\ false\:positive: IoU < \varepsilon
        \end{split}
\end{equation}

\noindent where $X_{W}$ and $X_{G}$ are spotted intervals and ME ground truth, respectively, and $\varepsilon$ is set as $0.5$.

In 2017, Tran et al. \cite{tran2017sliding} also proposed the first version of the ME spotting benchmark to standardize the performance evaluation of the ME detection task. Based on a the similar protocol for object detection, they utilized the same true positive detection as in equation \ref{overlapposneg} and plotted the Detection Error Tradeoff (DET) curve to compare spotters. While other works have also applied an interval-based evaluation similar to Li et at. \KHR{\cite{li2019spotting}}, they all proposed using another performance metric, such as the F1-score~\cite{tran2019,xu2016microexpression,li2020megc2020,see2019megc}.

In contrast to the above studies, several works follow the spotting of the apex frame location  \cite{yan2013fast,liong2015automatic} rather than detecting ME intervals. To evaluate the performance of these methods, such studies selected the Mean Absolute Error (MAE) to compute how close the estimated apex frames are to the ground-truth apex frames  \cite{yan2013fast,liong2015automatic,oh2018survey}. 

When performing spotting on longer videos, Liong et al. \cite{liong2016automatic} introduced another measure called the Apex Spotting Rate (ASR), which calculates the success rate in spotting apex frames within a given onset and offset range in a long video. An apex frame is scored one if it is located between the onset and offset frames, and 0 otherwise. In another approach, Nistor et al. \cite{nistor2018micro} proposed a combination of metrics, namely intersection over union percentage, apex percentage, and intersection percentage, to determine the number of correct detected apex frames.

Since there are various evaluation protocols for ME spotting, inconsistencies do result when comparing the performance among existing methods. Additionally, several unresolved issues still persist when evaluating the different ME spotting methods using existing protocols. For example, if two methods have the same number of true positive detections, no metrics exist for determining what method is better in terms of the ME interval coverage. Therefore, it is necessary to design a standard evaluation protocol that makes a fair comparison of ME spotting methods.

\section{SMIC-E-Long database}

In this section, we describe our work in constructing a new dataset for ME spotting. We realize the urgent need to explore the performance of the ME spotting task in a realistic environment. Thus, creating a new challenging database for ME spotting, one which has longer videos and more complex facial behaviors, is reasonable for exploring the performance of existing ME spotting techniques.

\subsection{Data Acquisition}


Construction of a completely new ME database often entails much effort and requires that experts label the ME ground truth. The SMIC-E and CAS(ME)$^2$~\cite{smicdata,casme2} databases have been extended to enable ME spotting by adding nature frames to the before and after ME samples. However, the number of frames added to these datasets is still limited. When constructing the \KHR{SMIC-E-HS dataset} \cite{pfister2011recognising,smicdata}, a large number of neutral frames remain from the recording step, which are captured by the high-speed camera. By adding these remaining frames from the previous recording, the new dataset might contain many challenging cases, such as head movements, eye blinking, and regular facial expressions, which can increase potential false detections. Thus, we decided to extend the \KHR{SMIC-E-HS dataset} by creating a more challenging spotting dataset, namely SMIC-E-Long.


In the beginning, we added $2000$ to $3000$ nature frames (approximately 20 seconds) before and after each ME sample to create longer videos (approximately 22 seconds per video). During this process, we combined several ME samples from the original SMIC database into a long video containing multiple ME samples. We also selected a few clips (around 20 seconds per video) without ME samples but containing regular facial expressions that might appear in real ME spotting situations.


Finally, we completed our new dataset with 162 long videos. Table~\ref{tab:table_compare} provides statistics from our new dataset compared with existing ME spotting datasets. There are approximately $350000$ frames in 162 long videos. We have 132 long videos containing ME samples and 25 videos with multiple MEs. There are 167 ME samples and 16 subjects in our dataset. The resolution of each video frame is $640 \times 480$. 
\KHR{The Frame-per-second (FPS) of the new SMIC-E-Long dataset is $100$, the same as the SMIC-E-HS dataset.}

\begin{table*}[!hbt]
\begin{center}
\caption{Comparison of the SMIC-E-Long and three other ME spotting datasets}
\label{tab:table_compare}
\resizebox{\columnwidth}{!}{
 \begin{tabular}{||c c c c c||} 
 \hline
 \textbf{Dataset} & \textbf{CAS(ME)\textsuperscript{2}} & \textbf{\KHR{SAMM-Long}} &  \textbf{SMIC-VIS-E} & \textbf{SMIC-E-Long} \\ [0.5ex] 
 \hline\hline
    Number of long videos & 97 & 147 & 76 & 162   \\
 \hline
  Number of subject & 22 & 32 & 8 & 16   \\
 \hline
    Videos with MEs & 32 & 79 & 71 & 132  \\
    \hline
    Resolution & 640 $\times$ 480 & 2040 $\times$ 1088 & 640 $\times$ 480 & 640 $\times$ 480  \\
    \hline
  ME samples & 57  & 159 & 71 & 166  \\
   \hline
    Average time \KHR{(seconds)}  & \KHR{87}  & \KHR{16} & 4  & 22  \\
    \hline
    \KHR{FPS}  & \KHR{30}   & \KHR{200}  & \KHR{25}   & \KHR{100}   \\

 \hline
 \hline
 \end{tabular}
 }
\end{center}
\end{table*}

\begin{figure}[ht]
    \centering
    \includegraphics[height=7cm,width=8cm]{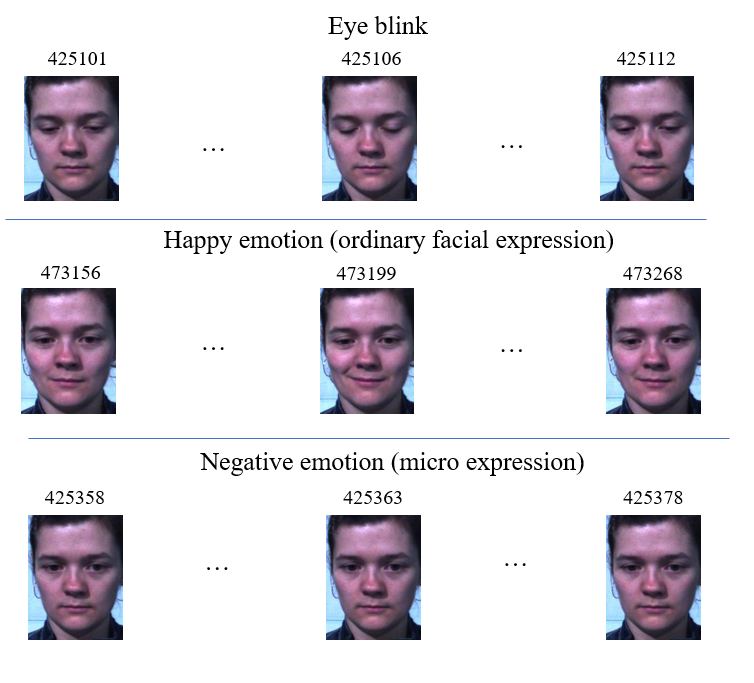}
    \caption{Example of three scenes in the new SMIC-E-Long dataset. The top case records an eye blink, 12 frames from frame index 425101 to frame index 425112. The middle case is the ordinary facial expression (happy emotion), which progresses from frame 473156 to frame 473268. The bottom case is an example of a negative emotion in the ME sample, which lasts from frame index 425358 to frame index 425378.}
    \centering
    \label{fig:outlier}
\end{figure}

Figure~\ref{fig:outlier} introduces other facial movements to our dataset. Two examples are presented: eye blinking and regular facial expression, both of which can easily be confused with MEs. As illustrated, eye blinking only takes 12 frames, so it is similar to the characteristics of the ME samples. These features make the SMIC-E-Long dataset challenging with respect to the ME spotting task.

\subsection{Face Preprocessing}

When first running the facial analysis system, we need to conduct the preprocessing step to align face images across video frames. This step is necessary to reduce differences in face shapes and changes caused by large face rotations throughout the video.


Generic face preprocessing includes four steps: (1) face detection and tracking to locate the face area throughout the video; (2) landmark detection to identify the specific landmark points on the face; (3) face registration to normalize the variations caused by head movements and different subjects; (4) face cropping to align the face-only area with a specific size.


\KHR{Existing studies have utilized various face alignment steps for differences in the final face size that affect the performance of the later spotting and recognition steps. For example, several methods have utilized Discriminate Response Map Fitting (DRMF)~\cite{liong2015automatic,han2018cfd}, while other studies have selected the Active Shape Model~\cite{li2017towards,xia2016spontaneous} to extract the landmark points.  In one study~\cite{li2019spotting}, the authors utilized Genfacetracker to extract the landmark points.}
This issue can result in an unfair comparison of different techniques. Therefore, we decided to carry out face alignment for our dataset and provide a preprocessed face set that can be used as standard input and make a fair comparison between methods.

\begin{figure}[ht]
    \centering
    \includegraphics[height=4cm,width=8cm]{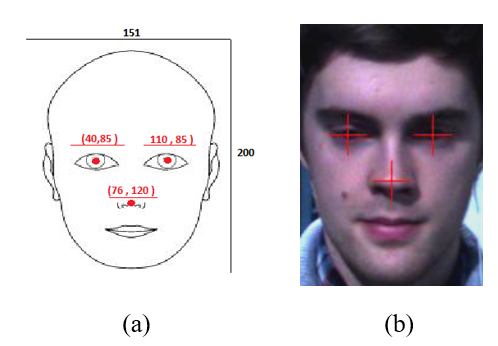}
    \caption{Illustration of a template for the face-preprocessing step. (a) The designed template when we carried out face alignment. (b) Example of an SMIC frame when applying face alignment, with the red crosses corresponding to three face registration points.}
    \centering
    \label{fig:template}
\end{figure}



We first located three landmark points in the first frame. Then, we applied face registration using the Local Weighted Mean method~\cite{goshtasby1988image}. The face area was cropped using the designed template. Figure~\ref{fig:template} illustrates the designed template for face alignment. It includes three landmark points and the final face size. After detecting the landmark points in one frame, we then utilized the landmark locations in the next consecutive $M = 30$ frames for the face alignment since the location of landmark points in short duration frames changes very little. For the frames with large head movements, we reduced the value of $M$ and conducted face alignment of the processing video once again to ensure the quality of the face alignment.



\section{Spotting Methods}
\label{spotting_method}
In this section, we select various ME spotting methods from unsupervised and supervised learning approaches to provide the baseline results.

\subsection{Unsupervised learning methods}

\subsubsection{$LBP-X^{2}$-distance method }

First, we implemented the method devised by Li et al. \cite{li2017towards}. This study provides the first baseline results for ME spotting. With this method, a scanning-window with size $L$ is stridden over a video sequence. In each position, we extracted LBP features from the $6 \times 6$ spatial block division in the first frame (HF), tail frame (TF), and center frame (CF) of the scanning window. Then, the distance between CF and the average value of TF and HF is computed based on the Chi-Square distance. Finally, the threshold method is applied to return the location of the apex frame of a detected ME. Details of this method are described on \cite{li2017towards}.

\subsubsection{Main Directional Maximal Difference Analysis for spotting}

The Main Directional Maximal Difference Analysis ($MDMD$) method was proposed by Wang et al.~\cite{wang2016main,wang2017main}. This method also utilizes a similar approach as the $LBP-X^{2}$-distance method~\cite{li2017towards} by using a scanning window and block-based division. $MDMD$ uses the magnitude of maximal difference in the main direction of optical flow as a feature for spotting MEs.

\subsubsection{Landmarks-based method}

We re-implemented the method from~\cite{beh2019micro} to explore the performance of spotting ME samples in longer videos by utilizing geometric features based on landmark points. With this method, specific landmark points on the eyebrows and around the mouth area are used to calculate the Euclidean distance ratio. A sliding window is then scanned across video frames to compute the change between the currently processing frame and the reference frame. A threshold method is used to decide what frames show MEs in the scanning window. Details of this method are described in~\cite{beh2019micro}.

\subsection{Supervised learning method}

\subsubsection{Spatial-temporal feature Method}

We also utilized the method proposed by Tran~\cite{tran2017sliding}. First, the ME samples are classified for each frame in a video sequence, and then non-maximal suppression is applied to merge the multiple detected samples. Spatial-temporal features, which are widely applied in ME and facial expression analysis, are extracted at each frame for the classification:

\begin{itemize}
    \item Histogram of Oriented Gradient for Three Orthogonal Planes ($HOG-TOP$). This feature has been extended from HOG to three dimensions to calculate the oriented gradients on three orthogonal planes for modeling the dynamic texture in a video sequence.
	\item Histogram of Image Gradient Orientation for Three Orthogonal Planes ($HIGO-TOP$). Histogram of Image Gradient Orientation ($HIGO$) is the degraded variant of HOG. It ignores the magnitude and counts the responses of the histogram bins.
	\item Local Binary Pattern for Three Orthogonal Planes ($LBP-TOP$). This feature was introduced in several studies~\cite{zhao2007dynamic,hong2016lbp}  as an extension of LBP for analyzing dynamic texture. It has been widely utilized as a feature for facial-expression analysis.

\end{itemize}

We utilized SVM to distinguish between the micro and non-micro samples. The details of this method can be found in~\cite{tran2017sliding}.

\subsubsection{CNN-based method}
\begin{figure}[ht]
    \centering
    \includegraphics[width=8cm]{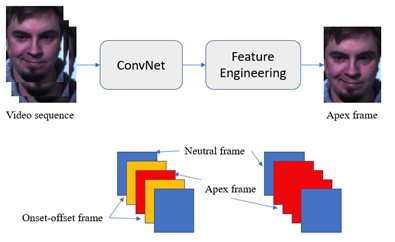}
    \centering
    \caption{Overview of the ME spotting method based on CNN architectures. First, CNN models are trained to determine the neutral frame and apex frame. Then, the trained models are utilized to extract features from each frame of the video. Finally, feature engineering techniques are proposed to return the final results.}
    \label{fig:convnet}
\end{figure}


Among deep learning structures, CNN has become popular since it offers good performance in many studies in the computer vision field. For ME spotting, we also applied CNN as a baseline for our dataset. The proposed method was developed based on the research presented in~\cite{zhang2018smeconvnet}.


As illustrated in figure~\ref{fig:convnet}, this method consists of two main components. The early step is a CNN model to distinguish between the apex and non-apex frame. We considered frames ranging from onset to offset as the apex frame to increase the number of apex frames for training. In contrast to the original work~\cite{zhang2018smeconvnet}, we selected two state-of-the-art CNN architectures for training the model: $VGG16$~\cite{simonyan2014very} and $ResNet50$~\cite{he2016deep}. We selected these two architectures because they have been considered the state-of-the-art image classification methods.

\begin{figure}[ht]
    \centering
    \includegraphics[width=8cm]{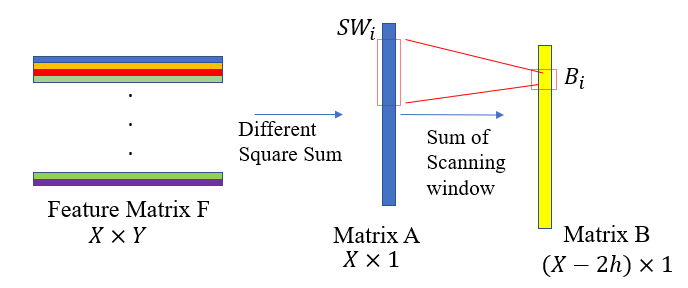}
    \centering
    \caption{Illustration of the feature engineering step from CNN-based spotting ($X$ is the number of frames in one long video, while $2h$ is the size of $SW_{i}$.).}
    \label{fig:fea_eng}
\end{figure}

The later step involves feature engineering, which is used to combine the nearby detections. The features extracted from each frame of the video sequence are concatenated to construct feature matrix $F$ with dimension $X \times Y$, where $X$ is the number of the frame and $Y$ is the length of the features in the last layer of the CNN models. Each row of feature matrix $F$ corresponds to a frame in the long video. After constructing $F$, we calculated the difference square sum based on the values of each row compared with the first row. The results were concatenated to build matrix $A$. Then, we slid a scanning-window across the positions of matrix $A$ at size $2 \times h$ to sum the values inside the window (the result was returned as matrix $B$). The maximum value from the sum was considered the apex frame. Figure~\ref{fig:fea_eng} illustrates the feature engineering step.

\KHR{After spotting the apex frame, we applied non-maximal suppression to merge the nearby detected apex frame. Specifically, we selected the highest apex score. Then, we considered the $h$ frames after and before the detected apex frame as the ME frames to compare with the spotting method using interval-based techniques.}

\subsubsection{LSTM-based method}

\begin{figure}[ht]
    \centering
    \includegraphics[width=8cm]{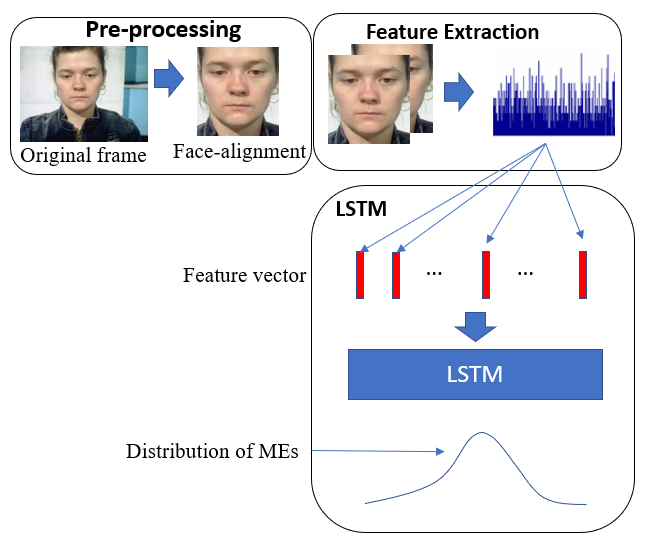}
    \centering
    \caption{Overview of LSTM-based method for ME spotting. First, we extracted a spatial-temporal feature at each temporal position $i$ of a video sequence. Each feature vector takes the information of $L$-consecutive frames from temporal location $i$ to $i + L$. After extracting the features, we slid a scanning window across the video to create a sequence of features for the LSTM model. In each scanning window, we predicted the score of a candidate ME sample.}
    \label{fig:lstm}
\end{figure}

Another baseline method for ME spotting is the sequence-based learning approach proposed by Tran et al.~\cite{tran2019}. This method includes two main steps: feature extraction and apex frame detection based on a deep sequence model. Figure~\ref{fig:lstm} provides an overview of the method.


To construct the sequence of spatial-temporal feature vectors, we slid a scanning window across all temporal positions of a video. In each position, for example at frame $i$, we extracted features in the sequence from frame $i$ to $(i + L)$. By using this strategy, we considered each position to be a ME candidate. In one study~\cite{tran2017sliding}, the authors proposed the continuous ground truth score for each sample. However, we ignored these scores and considered one sample as either
ME or non-ME, meaning that if one position is considered an ME sample, the ground truth is set as 1, otherwise as 0. For the spatial-temporal features, we utilized two kinds of features: $HIGO-TOP$ and $HOG-TOP$~\cite{li2017towards}.

For the apex frame detection, we employed the Long short-term memory (LSTM) network, which is a special kind of Recurrent Neural Network (RNN). It was introduced by Hochreiter~\cite{hochreiter1997long}, and it has since been refined and popularized in many studies, especially for sequence learning. The power of the LSTM network in learning and modeling sequence data is useful for estimating an ME’s position in the video sequence.


The idea behind using LSTM in ME spotting is to slide a scanning window across the video sequence. Each scanning window contains $M$ spatial-temporal features corresponding to $M$ temporal positions. The constructed network inputs $M$ spatial-temporal features and predicts $M$ values representing the score of MEs at each temporal position. The feature at positions $(i^{*})$ with the highest score inside a scanning window and that is larger than the threshold is considered an ME sample. The specific apex frame is determined by the middle location $\frac{i^{*} + (i^{*}+L) }{2}$ of the feature.


When dealing with ME spotting in the long videos, most of the methods utilize the scanning window to spot ME samples. Therefore, after predicting the ME score in each frame, we needed to combine the nearby detected samples using the same feature engineering strategy as the CNN-based spotting method. \KHR{Additionally, to spot interval in the ME sample, $\frac{L}{2}$ frames occurring after and before the apex frame are considered ME frames.}


\section{Proposed evaluation protocol}
\label{evaluation}

As mentioned in the related work section, several existing studies have proposed various performance evaluation protocols. It is difficult to say which one is better for ME spotting. To standardize the performance comparison, we recommend using a new set of protocols that is more generalized and suitable for the evaluation of ME spotting. Our proposed protocols will be described in the next sub-sections: (1) the experimental setup for splitting the data into the training and testing sets and (2) the metric for evaluating the performance of spotting techniques.

\subsection{Training and testing set} \label{evaluation_split}

To split data into training and testing sets, we selected Leave-one-subject-out (LOSO) cross validation. LOSO is a training setup widely utilized for ME analysis and facial expression recognition problems~\cite{li2017towards,tran2019,jain2019classification}. Particularly, with this method the videos and samples belonging to one subject (participant) are kept for testing, while the remaining samples are used for training.

\subsection{Performance Evaluation}

For standardizing the performance evaluation of ME spotting methods, we propose two new ME movement, spotting-based evaluation metrics: sample-based and frame-based metrics. The evaluation using single apex frame detection, which is usually employed in previously proposed evaluation protocols, is not considered in our protocol. We argue that the information from a single detected apex frame is not sufficient for the following ME recognition step since most ME recognition methods often require the correct extraction of the ME interval. Therefore, we want to focus on the evaluation metric that identifies the correctly detected ME interval as well.

\subsubsection{Sample-based evaluation}

In this section, the sample-based evaluation metric for the detected ME samples is introduced. First, we discuss how to decide whether one spotted interval is a true positive or false positive. With many ME spotting methods, the Intersection over Union (IoU) is a common approach that tackles the ME spotting as an object detection problem. This method is applied to the ME spotting task by considering the overlap between ME samples and spotted intervals (it is similar to comparing the boundary box of detection with ground truth in an object detection problem~\cite{dollar2009pedestrian}). The decision condition is shown in equation~\ref{overlapposneg}. The value of $\varepsilon$ is set as $0.5$ in most of the studies. It is arbitrary but reasonable. Then, we utilized the same evaluation metric as Li et al.~\cite{li2019spotting} to compare the spotting methods.


To provide a more informative comparison, we also present the results of several methods following the DET curve evaluation protocol, utilizing the research of Tran et al.~\cite{tran2017sliding}. Following Tran et al.~\cite{tran2017sliding}, we created a Detection Tradeoff Error (DET) curve to plot the False-Positive Per-video (FPPV) versus miss rate values. The values FPPV and Miss rate are calculated using equation~\ref{fppv_cal}:

\KHR{
\begin{equation}
    \label{fppv_cal}
    \begin{split}
    FPPV_{OverAll} = \frac{\sum_{j=1}^{S} FP_{j}}{V} \\
    MissRate_{OverAll} = 1 - \frac{\sum_{j=1}^{S} TP_{j}}{N^{+}}
    \end{split}
\end{equation}
}


\noindent where $N^{+}$ is the number of ME samples in our dataset, $V=162$ is the number of long videos in our database, \KHR{and $S=16$ is the number of subjects.}

\begin{figure}[ht]
      \centering
  \includegraphics[scale=0.5]{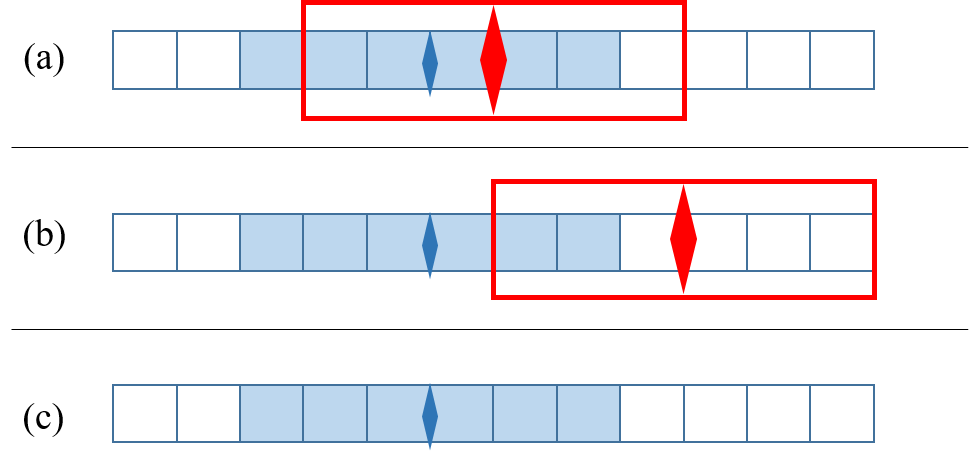}
  \caption{Illustration of detected sample and ground truth, with (a) being an example of one hit, (b) being an example of one false positive, and (c) being an example of one miss with no detected sample.}
  \label{fig:sample_define}
\end{figure}

Even still, the use of IoU has a drawback caused by the different lengths of the ME samples in our database, which range from $10$ to $51$ frames. Therefore, several ME samples can be missed when using a fixed length detected window, which has usually been implemented in previous ME spotting approaches~\cite{xia2016spontaneous, tran2019}. For example, if we set the detected window length at $35$, ME samples having a size of $11$ to $17$ will be missed in the evaluation based on IoU having a threshold of 0.5. Therefore, we defined a new evaluation metric for the detected sample as follows. In the first step, we defined the three values for comparison as ``Hit'' (TP), ``Miss'' (FN), and ``False'' (FP). ``Hit'' is counted when a ME ground truth in the center frame ($C\_gt$) satisfies the condition for the nearest detected window:  $ |C\_w - C\_gt | \leq 0.5 \times L\_gt $, where $C\_w$ ,$C\_gt$ are the center location of the detected window and ground truth, respectively, and $L\_gt$ is the length of the detected ground truth. Otherwise, the un-spotted ground truth is counted as one ``Miss.'' ``False” is counted when one detected window has the nearest ground truth satisfying the condition $ |C\_w - C\_gt | > 0.5 \times L\_gt $. Figure~\ref{fig:sample_define} provides an illustration of the TP sample, FP sample, and FN sample.

After calculating the three values TP, FP, and FN, we needed to consider two issues regarding the F-measure values. The first one has to do with missing the ME in the ground truth or no ME samples being detected. This problem causes a division by zero when computing the recall and precision. The second issue has to do with the difference in the number of videos from each subject. Several subjects only have 4 to 6 videos. If we calculate the average value of 16 folds (corresponding to 16 subjects), it will cause an imbalance in the contribution of each subject to the final performance. To resolve these issues, at every LOSO testing fold we integrated all tested videos into a single long video. After that, the values of TP, FP, and FN obtained in each testing phase were added up to calculate the (overall) precision, recall, and F1 score of the entire database. Then, we calculated the precision, recall, and F1-score using equation~\ref{Fscore_macro}:

\begin{equation}
    \label{Fscore_macro}
    \begin{split}
    Precision = \frac{TP_{all}}{TP_{all} + FP_{all} }
    \\
    Recall = \frac{TP_{all}}{TP_{all} + FN_{all} }
    \\
    F1-score = \frac{2*Precision*Recall}{Precision + Recall}
    \end{split}
\end{equation}


\noindent where $TP_{all} = \sum_{i=1}^{V} TP_{i} $ , $FP_{all} = \sum_{i=1}^{V} FP_{i} $, and $FN_{all} = \sum_{i=1}^{V} FN_{i} $ are the cumulative values of TP, FP, and FN at each testing phase, respectively. $V$ is the number of videos, while $TP_{i}$, $FP_{i}$ and $FN_{i}$ are the number of true positive, false positive, and false negative frames, respectively, in $i^{th}$ videos.

\subsubsection{Frame-based evaluation}

\begin{figure}[ht]
      \centering
  \includegraphics[scale=0.5]{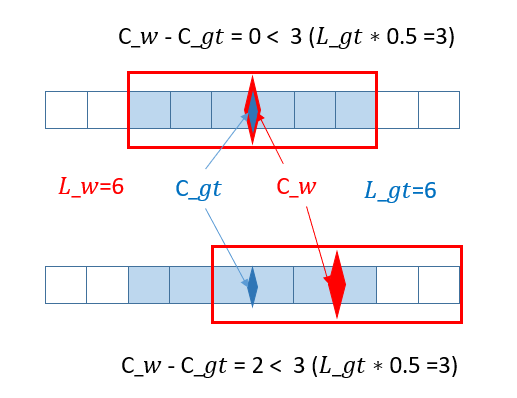}
  \caption{The limitations of using only sample-based evaluations. Both spotters are true positive samples, but the sample-based method cannot conclude that the top spotter is better than the bottom spotter.}
  \label{fig:hit_fail}
\end{figure}


In the previous sub-section, we described the performance evaluation based on the detected ME samples. A sample-based evaluation is a common approach when evaluating ME spotting methods. It is quite similar to evaluating object detection problems. However, ME spotting is a different topic than object detection. If we consider building an automatic system to detect and recognize MEs in longer videos, the quality analysis of the detected samples should be explored. A bad ME sample could have too many neutral frames or miss ME frames for the next recognition task. Several recognition studies have specified that spotted ME samples should be detected correctly (from onset - apex - offset)~\cite{li2015reading,van2019capsulenet,wang2014micro,huang2016spontaneous}. Therefore, missing ME frames or any redundancy in the non-micro frames in the spotting results can greatly affect the recognition performance.


To address this issue, it is important to consider the frame-based evaluation, which focuses on the correctness
of spotted frames in each detected true positive sample. We complemented this with a metric called frame-based accuracy \emph{F}. The frame-based accuracy \emph{F} metric can be defined as the mean deviation of ``lengths'' and ``centers'' in all correctly detected (TP) samples. The calculation of \emph{F} is defined in equation~\ref{frame_rate}:

\begin{equation}
    \label{frame_rate}
    F = \frac{1}{n} \sum_{i=1}^{n}  \frac{((|C\_w_{i} - C\_gt_{i}|)  + (|L\_w_{i} - L\_gt_{i}|))}{2*L\_gt_{i}} 
\end{equation}


\noindent where $C\_w_{i}$ and $L\_w_{i}$ are the center frame and window size of the $i^{th}$ true positive sample, $C\_gt_{i}$ and $L\_gt_{i}$ are the center frame and length of the ME ground truth corresponding to the $i^{th}$ true positive sample, respectively, and $n$ is the number of true positive samples. Figure~\ref{fig:hit_fail} illustrates the effectiveness of the \emph{F} metric. Based on this figure, we can conclude that the top spotter in the detected window is better than the bottom detector. A smaller \emph{F} value means more accurately detected ME locations. When considering an extreme case, if all TP cases perfectly match with the ground truth, that is, the top spotter in figure~\ref{fig:hit_fail}, then \emph{F} is $0$.


\KHR{However, the proposed formula in equation~\ref{frame_rate} can only be used to estimate interval-based ME spotting methods and cannot be applied to single apex frame methods. For single apex frame spotting, we propose using equation~\ref{apex_frame} by removing the factor $|L_{w} - L_{gt}|$. In this way, the new formula can be used at the same time for single apex frame spotting as well as interval-based ME spotting:}

\KHR{
\begin{equation}
    \label{apex_frame}
    F = \frac{1}{n} \sum_{i=1}^{n}  \frac{(|C\_w_{i} - C\_gt_{i}|)}{L\_gt_{i}} 
\end{equation}
}

\section{Experiments}

\subsection{Implementation}




We describe here the implementation and training details of our selected baseline methods. The required development tools and the selection of important parameters will also be discussed.

First, to determine the value for $L$, the length of detected ME samples, we calculated the mean value of the ME sample size. Since the mean value was $34$, we set the detected window size at $L = 35$. For ME spotting methods that return only the location of apex frames, such as the $MDMD$ method, $\frac{L}{2}$ frames after and before the spotted apex frame
were considered as being inside the ME samples.


Next, we explain how each method was implemented. First, we considered unsupervised learning methods. With respect to the $LBP-X^{2}$-distance method~\cite{li2017towards}, the authors did not conduct any spotting on the aligned faces. However, we conducted our experiment on pre-processed faces. For extracting the LBP features, we utilized the \textbf{scikit-image} with the default setup~\cite{van2014scikit}.


With method $MDMD$, we utilized the source code provided by the authors. In the original source code, the $MDMD$ method outputs only the locations of the apex frames. Since some detected apex frames can be quite close to each other, overlaps may occur in the extracted ME samples. To guarantee consistency with other baseline methods, we employed Non-Maximal Suppression (NMS) to keep only one detected ME sample in every $L$ continuous frame. We will present the evaluation results of both the original $MDMD$ method and the $MDMD$ method with NMS ($MDMD~NMS$) in our experiments.


We re-implemented the landmark-based method by utilizing the DLIB toolbox to extract $68$ landmark points for calculating the ratio between specific points. The window size of the processing landmark points was also set as $L$.



Regarding the spatial-temporal feature methods, we utilized the MATLAB tool provided by the authors to extract three features, $LBP-TOP$, $HIGO-TOP$, and  $HOG-TOP$~\cite{dollar2009pedestrian}. Following the research done by Tran et al.~\cite{tran2017sliding}, we only selected one set of parameters: block division $8 \times 8 \times 4 $, overlap $0.2$, with the number of bins being $8$. The length of images sequence for spatial-temporal feature extraction was set as $L$, and we employed the \textbf{scikit-learn} toolbox to train the model. \KHR{With these methods, we had expected to scale the longer video into particular lengths. Several techniques~\cite{hong2019characterizing,zhou2011towards,zhou2013compact,peng2019boost} can be used to interpolate video frames into videos having the same length. We selected the Temporal Interpolation Model~\cite{zhou2011towards} to scale the video with scale factors $1$, $0.75$ and $0.5$}. We aimed to return the adaptive size of each detected window. Three spatial-temporal feature methods were denoted as $HIGO-TOP-SVM$, $HOG-TOP-SVM$, and $LBP-TOP-SVM$ in the experiment results.


For the CNN-based methods, $VGG16$ and $ResNet50$ architectures were employed to train the neutral and apex frame classifier. We set the learning rate at $0.001$, the number of epochs at $50$, and the optimizer as Stochastic Gradient
Descent. \KHR{The value $h$ was set as $\frac{L}{2} = 17$.} We denoted these two methods as $VGG16$ and $ResNet50$ in the reported
table. For the LSTM method, we set the learning rate at $0.001$ and used the Adam configuration for the optimizer.
The spatial-temporal features ($HIGO-TOP$ and $HOG-TOP$) were extracted in the same way as in the spatial- temporal feature method. The lengths of the detected ME samples and the LSTM sequence were set as $L$ and $M=50$, respectively. \KHR{We selected $M=50$ to represent half of a second.} These methods are indicated as $HIGO-TOP-LSTM$ and $HOG-TOP-LSTM$. Deep learning architectures were constructed using the Keras framework and trained on the GPU Tesla K80 server.


\KHR{Additionally, we also conducted our experiments using three public datasets, SMIC-VIS-E, CAS(ME)$^{2}$, and SAMM- Long. We set the spotting method parameters for each dataset as follows: for SMIC-VIS-E with $FPS=25$, we set them as $L=9$ and $M=25$; for CAS(ME)$^{2}$ with $FPS=30$, we set $L$ and $M$ as $13$ and $25$, respectively; for SAMM-long, we set them as $FPS = 200$. Thus, we set the parameters as $L = 65$ and $M = 50$. The other hyperparameters were set the same as with the SMIC-E-long experiments.}

\subsection{Results} \label{results}

In this section, we present the baseline results based on our proposed evaluation protocols. We used LOSO to split the dataset into training and testing sets, as discussed in section~\ref{evaluation_split}. Details of each set of experiments are described in the next following sub-sections.

\subsubsection{Comparison on existing protocols}

\begin{figure}[ht]
      \centering
  \includegraphics[scale=0.5]{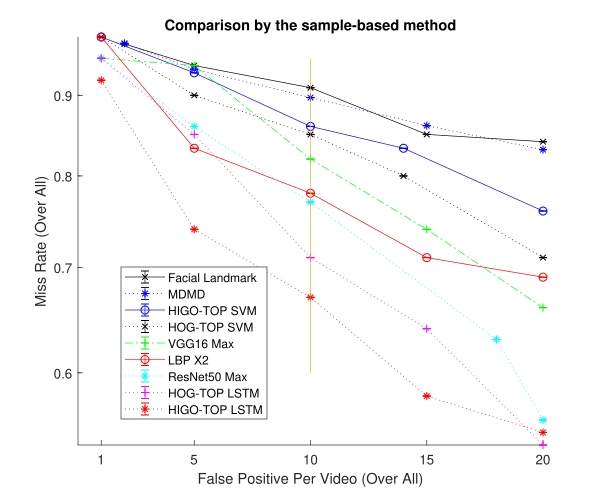}
   \centering
  \caption{Performance of detectors based on interval domain (lower is better).}
  \label{fig:det_interval}
\end{figure}

In this set of experiments, we evaluated the baseline methods based on the existing evaluation metrics: precision, recall, and F1-score~\cite{li2019spotting}. The results are displayed in table~\ref{tab:table_interval} and table~\ref{tab:compare_data}. The performances of the DET curve are displayed in figure~\ref{fig:det_interval}.


\begin{table*}[!hbt]
\begin{center}
\caption{\KHR{Experiment results for the SMIC-E-Long dataset based on the existing protocols (F1 measure with IoU). The methods are categorized into two approaches: supervised learning and unsupervised learning.}}
\label{tab:table_interval}
 \begin{tabular}{ c c c c c c c c } 
 \hline
 \textbf{Approach} & \textbf{Method}  & \textbf{TP}  & \textbf{FP} & \textbf{FN} &  \textbf{Precision} $\uparrow$  & \textbf{Recall} $\uparrow$ & \textbf{F1-score} $\uparrow$ \\ [0.4ex] 
 \hline\hline
    \multirow{4}{*}{\parbox{2cm}{\centering Unsupervised Learning}} & Facial landmark  &  15 & 846 & 151 & 0.01745  & 0.0903 & 0.0292  \\
    & MDMD  &  13 & 2934 & 153 & 0.0044  & 0.0783 & 0.0081 \\
    & MDMD~NMS  &  11 & 644 & 155 & 0.0168  & 0.0662 & 0.0268 \\
    & $LBP-X^{2}$  &  21 & 443 & 145 & 0.0452  & 0.1265 & 0.0666  \\
    \hline
    \multirow{7}{*}{\parbox{2cm}{\centering Supervised Learning}} & LBP-TOP-SVM$^*$  & 6 & 218 & 160 & 0.0267  & 0.0361 & 0.0307 \\
    & HIGO-TOP-SVM$^*$  & 14 & 857 & 152 & 0.0505 & 0.0602 & 0.0270 \\
    &  HOG-TOP-SVM$^*$  & 11 & 416 & 155 & 0.0922 & 0.1144 & 0.0371 \\
    & HOG-TOP-LSTM  & 55 & 1833 & 111 & 0.0291 & 0.3313 & 0.0535  \\
    & HIGO-TOP-LSTM  & 31 & 545 & 135 & 0.0538  & 0.1867 & \textbf{0.0835} \\
    & ResNet50 & 11 & 234 & 155 & 0.0449 & 0.0662 & 0.0535   \\
    & VGG16  & 9 & 428 &  157 & 0.0205 &  0.0542 & 0.0298  \\
    \hline
 \hline
 \end{tabular}
\end{center}
\footnotesize{$^*$ refers to handcrafted feature techniques in the supervised learning approach.}
\end{table*}



Table~\ref{tab:table_interval} shows that $HIGO-TOP-LSTM$ proved to be the best method based on the F1-score: $0.0835$. $LBP-X^{2}$ method yielded an F1-score of $0.0666$, which was second best. In figure~\ref{fig:det_interval}, we selected $FPPV = 10$ as the reference point to analyze the miss rate values (lower is better). The $HIGO-TOP-LSTM$ method was the best ME spotting method, with a miss rate of $0.681$. At the same reference points, another LSTM-based method ($HOG-TOP-LSTM$) ranked in second place with a miss rate of $0.721$.

With respect to the SMIC-E-Long dataset, another reason for the low F1-score values was the small number of TP samples. In terms of the final output of each spotter, we found that several ME samples could not be detected because the size of the ME samples was too small. As mentioned in section~\ref{evaluation}, ME samples having a size of less than $15$ cannot be considered TP if the length of the detection window is set higher than $30$. Additionally, in the case of MEs
with a very low intensity, the ME spotters only approximately located the apex frame inside the onset and offset of the ME interval, and hence, reduced the IoU value. Therefore, it resulted in missing ME samples when utilizing the IoU to decide whether one spotted ME interval was a TP or FP. In summary, we realized that the evaluation method for ME spotting that employs IoU has a limitation caused by the diverse sizes of the ME samples.

\subsubsection{Comparison on proposed  protocol}



The experimental results for the selected baseline methods with the proposed protocol are shown in table~\ref{tab:new_table}. Note that the performance of the baseline methods increased. For example, in the results for the $HIGO-TOP-SVM$ method, the number of TP samples increased from $14$ to $24$. Consequently, the F1-score also improved from $0.0270$ to $0.0604$. When analyzing the final output from each spotter, some ME samples, as discussed in the last experiment, were not correctly spotted when utilizing the IoU evaluation. In contrast, many of those samples are now considered TP samples in our proposed protocol because we did not consider the length of the detected samples in the evaluation. We only considered the center frame location of the detected window and the ME samples. Therefore, the missing TP ME samples that resulted from applying the IoU evaluation has now been alleviated, which is fairer and more reasonable than existing evaluation metrics.


Regarding the number of TP samples, we found that the $HOG-TOP-LSTM$ method was the best, with $TP = 66$, and the $HIGO-TOP-LSTM$ method was second best, with $TP = 36$. The other methods returned a TP number of less than $30$. When comparing them by F1-score, the best method was the $HIGO-TOP-LSTM$ method, with an F1 score of 0.0970. The second and third best methods were  $LBP-X^{2}$ and $HOG-TOP-SVM$, respectively.


\KHR{With respect to the frame-based evaluation, table~\ref{tab:new_table} reports the F values of each baseline method, with the $LBP-X^{2}$ method returning the best $F$ value, $0.1483$, followed by the $ResNet50$ method, with an $F$ value of $0.1495$. The third best method was the $HOG-TOP-LSTM$ method, for which $F = 0.1922$.}


\begin{table*}[!hbt]
\begin{center}
\caption{\KHR{Experiment results following our proposed protocols on the SMIC-E-Long database. The F1-score$^{\Delta}$ was calculated using equation~\ref{Fscore_macro}. The methods were categorized into two approaches: supervised learning and unsupervised learning. }}
\label{tab:new_table}
 \begin{tabular}{c c c c c c c c c c} 
 \hline
 \textbf{Approach} & \textbf{Method} & \textbf{TP}  & \textbf{FP} & \textbf{FN} &  \textbf{Precision}  & \textbf{Recall} & \textbf{F1-score$^{\Delta}$} $\uparrow$ & \textbf{F} $\downarrow$ \\ [0.4ex] 
 \hline\hline
     \multirow{4}{*}{\parbox{2cm}{\centering Unsupervised Learning}} & Facial~landmark  &  20 & 841  & 146 & 0.0232  & 0.1204 & 0.0389 & 0.2236 \\
     & MDMD &  24 & 2900 & 142 & 0.0082  & 0.1445 & 0.0155 & 0.7411 \\
    & MDMD~NMS  &  19 & 636 & 147 & 0.0290  & 0.1144 & 0.0462 & 0.2777  \\
    & LBP-X$^{2}$  &  26 & 438 & 140 & 0.0560  & 0.1566 & 0.0825 & \textbf{0.1483} \\
    \hline
    \multirow{7}{*}{\parbox{2cm}{\centering Supervised Learning}} & LBP-TOP-SVM$^*$ & 10 & 214 & 156 & 0.0446  & 0.0602 & 0.0512 & 0.2179 \\
     & HIGO-TOP-SVM$^*$  & 24  & 847 & 146 & 0.0555  & 0.0662 & 0.0604 & 0.2444 \\
     & HOG-TOP-SVM$^*$  & 21 & 406 & 145 & 0.0491 & 0.1265 & 0.0708 & 0.2502 \\
    & HOG-TOP-LSTM  & 66 & 1930 & 96 & 0.0350 & 0.4216 & 0.0642 & 0.1922 \\
    & HIGO-TOP-LSTM  & 36  & 540 & 130 & 0.0620  & 0.2108 & \textbf{0.0970} & 0.2337 \\
    & ResNet50 & 12 & 233 & 154 & 0.0489 & 0.0722 & 0.0583 & 0.1495 \\
    & VGG16  & 10 & 427 & 156 & 0.0228 & 0.0602 & 0.0441 & 0.2028 \\
    \hline
    \hline
 \end{tabular}
\end{center}
\footnotesize{$^*$ refers to handcrafted feature techniques in the supervised learning approach.}\\
\end{table*}



To demonstrate the challenge inherent to our dataset, we provide the experimental results for the three existing datasets: CAS(ME)$^2$, SMIC-VIS-E, and SAMM-long~\cite{casme2,smicdata,yap2019samm}. As presented in table~\ref{tab:compare_data}, the performance of the baseline methods in the SMIC-VIS-E dataset was better than the three other databases. Almost all the F1-score values in the SMIC-VIS-E dataset were greater than $0.3$, while the performance of the SMIC-E-Long dataset showing poorer results, with most of the methods having an F1 score values of less than $0.1$. As shown in table~\ref{tab:table_interval}, the number of FP samples was quite high because all the baseline methods had difficulty in discriminating between ME samples and other facial movements, which introduced additional challenges for our new dataset.

\begin{table*}[!hbt]
\caption{\KHR{Comparison between our SMIC-E-Long and three existing datasets evaluated by the new metric. The methods are categorized into two approaches: supervised learning and unsupervised learning. (*) means that handcrafted feature techniques in the supervised learning approach. Equation \ref{Fscore_macro} calculates  F1-score values on this table.}}
\begin{center}
\label{tab:compare_data}
 \begin{tabular}{c c c c c c} 
 \hline
 \textbf{Approach} & \textbf{Method} & \textbf{CAS(ME)$^{2}$}  & \textbf{SMIC-VIS-E} & \textbf{SAMM-Long} & \textbf{SMIC-E-Long} \\ [0.5ex] 

 \hline\hline
    \multirow{2}{*}{Unsupervised Learning} & MDMD~NMS  &  0.0124 & 0.1544 &  0.0521 & 0.0462  \\
    & $LBP-X^{2}$  &  0.0125 & 0.1496 & 0.0630 & 0.0825  \\
    \hline
      \multirow{4}{*}{Supervised Learning} & HOG-TOP-SVM$^{*}$  & 0.0068 & 0.3175 & 0.0571 & 0.0604   \\

   & HIGO-TOP-SVM$^{*}$  & 0.0068  & 0.3432 & 0.0553 & 0.0708  \\

    & HOG-TOP-LSTM  & 0.0092 & 0.6221 & 0.06006 & 0.0642   \\

    & HIGO-TOP-LSTM  & 0.0122  & 0.5532 & 0.0541 &  0.0970 \\
    \hline
 \hline
 \end{tabular}
\end{center}
\end{table*}






\subsubsection{Discussion}
In this sub-section, we analytically demonstrate the effectiveness of our selected spotting methods. Our analysis is split into five parts based on the following details.

\textbf{Unsupervised method vs. supervised method.} We first explored the effectiveness of unsupervised and supervised approaches. Based on the results of the F1-scores shown in table~\ref{tab:table_interval} and table~\ref{tab:new_table}, we can observe that both methods performed similarly:  the best method, though, was the  $HIGO-TOP-LSTM$  method (a supervised method), while the second best was the $LBP-X^{2}$ method (an unsupervised method). When considering the $FP$ values, most of the supervised methods proved better than the unsupervised methods. This issue can be explained by the fact that the supervised methods have more labeled data in training to distinguish ME samples from non-ME samples.


The ($LBP-X^{2}$) method did not utilize the training term but surpassed several machine learning techniques to rank second best. Unlike other methods, ($LBP-X^{2}$) employs the adaptive threshold for each video based on the motion values. Hence, the reasons for better performance may stem from the following: (1) the adaptive thresholds combined with the textural features were robust enough to reject the non-ME facial movements; (2) the facial actions were largely different across subjects in our new dataset. The latter can cause an overfitting issue in the supervised learning techniques, which decreases performance during testing.


\textbf{Handcrafted features vs. deep features.} Next, we explored the performance of supervised methods, which can be further split into two approaches: handcrafted feature learning and deep feature learning methods. Table~\ref{tab:new_table} interestingly shows that the use of LSTM is not always better than the traditional machine learning techniques in term of the F1-score ($HOG-TOP-SVM$  was better than $HOG-TOP-LSTM$, while $HIGO-TOP-LSTM$ achieved  better performance than $HIGO-TOP-SVM$). This issue can be explained by the difference between the $HIGO-TOP$ and $HOG-TOP$ features. Therefore, the correlation between spatial-temporal features and learning models is an interesting topic to explore in future research. Additionally, the number of ME samples for training is still not enough, while deep learning techniques often require a huge amount of training data.


Another deep learning technique in our baseline was the CNN-based method, which had lower performance since it is difficult to discriminate between micro frames and neutral frames when only using a single frame. As mentioned in the specification for the CNN-based ME spotting method, frames ranging from onset to offset in the ME samples are considered apex frames for training the image classification model. However, when observing a single frame, several extrinsic facial movements also displayed similar facial action units as ME samples. This issue becomes confusing in the image
classification model when discriminating between ME frames and frames containing other facial movements. Hence, CNN-based methods can detect the frames of other facial movements, such as ME apex frames. In general, it appears that the use of deep sequence learning for ME spotting is promising, but the number of ME samples needs to be increased to further improve the performance.

\textbf{Appearance features vs. geometric feature.} In our experiments, most methods aimed to spot ME samples based on appearance features: spatial-temporal feature (LSTM-based methods and SVM-based methods), optical flow ($MDMD$), texture analysis, and raw image. We only had one method that used geometric features (facial landmark points), but this method had lower results compared to the appearance-based approaches. The problem with the facial landmark method is that it is based on the rather simple idea of calculating the distance ratio between landmark points. Therefore, this method cannot detect several ME intervals with overly small movements. Especially, several ME cases only occurred in the eye regions, making them similar to eye blinking actions. These cases resulted in missing TP samples or the triggering of false detections.


\textbf{Frame-based performance.} \KHR{For the reported ``F'' values shown in table~\ref{tab:new_table}, the single-apex frame detector yielded the best two results, which can be explained by the fact that single-apex frame methods spot the center frame of the ME sample better than the interval spotters. In our experiments, the SVM-based methods carried out multi-scale detection to return the adaptive-size windows. However, the ``F'' values for these methods were still high. Additionally, the ``F'' values for the SVM-based methods were not any better than other methods with similar TP values. To address this issue, future research needs to consider what spotting techniques can provide output for the detected adaptive window or focus on the spotting of the onset and offset locations.}

\textbf{Databases comparison.} \KHR{In table~\ref{tab:compare_data}, we compare the new dataset with three existing datasets. Based on the results,
the dataset with short videos (SMIC-VIS-E) performed than the other three databases with long video sequences. The videos in the SMIC-VIS-E dataset only had ME frames and neutral frames, which is much less challenging with less false detections, while the other three datasets also contained normal facial expressions, eye blinking, and head movements. Successfully resolving the ME spotting task in long and natural videos is still a very challenging problem for future research.}

\section{Conclusion}
In this paper, we introduced a new challenging dataset for ME spotting. We also suggested a new set of evaluation protocols to standardize the comparison of ME spotting techniques and enable a fairer comparison. Finally, we explored various spotting algorithms ranging from handcrafted to deep learning methods to provide the baseline for future comparisons. 

Based on the experimental results, we identified several issues that can still be improved upon in future studies. First, the poor performance of deep learning techniques can be improved by applying data augmentation to increase the number of ME samples. We can employ better deep sequence models to analyze the spatial-temporal information. Cross-database evaluation needs to be conducted to explore the generalization capability of ME spotting techniques
in future research. Furthermore, the detection of the onset-offset frames should be included in follow-up studies to locate the ME samples correctly. Finally, the ME recognition will be considered to create a complete benchmark for the whole automatic ME analysis system.



\bibliographystyle{elsarticle-num} 
\bibliography{main}


\end{document}